\documentclass{article}
\usepackage{spconf,amsmath,epsfig,amssymb,graphicx,subfigure}

\title{Iterative Reweighted Algorithms for Sparse Signal Recovery with Temporally Correlated Source Vectors}
\name{Zhilin Zhang and Bhaskar D. Rao\thanks{The work was supported by NSF Grant CCF-0830612.}}
\address{Department of Electrical and Computer Engineering,\\
         University of California at San Diego, La Jolla, CA 92093-0407, USA \\
         \{z4zhang,brao\}@ucsd.edu}
\begin{document}
\ninept

\maketitle
\begin{abstract}
Iterative reweighted algorithms, as a class of algorithms for sparse signal recovery, have been found to have better performance than their non-reweighted counterparts. However, for solving the problem of multiple measurement vectors (MMVs), all the existing reweighted algorithms do not account for temporal correlation among source vectors and thus their performance degrades significantly in the presence of correlation. In this work we propose an iterative reweighted sparse Bayesian learning (SBL) algorithm exploiting the temporal correlation, and motivated by it, we propose a strategy to improve existing reweighted $\ell_2$ algorithms for the MMV problem, i.e. replacing their row norms with Mahalanobis distance measure. Simulations show that the proposed reweighted SBL algorithm has superior performance, and the proposed improvement strategy is effective for existing reweighted $\ell_2$ algorithms.
\end{abstract}
%

\section{Introduction}
\label{sec:intro}

The multiple measurement vector (MMV) model for sparse signal recovery is given by \cite{Cotter2005}
\begin{eqnarray}
\mathbf{Y}= \mathbf{\Phi} \mathbf{X} + \mathbf{V},
\label{equ:MMV basicmodel}
\end{eqnarray}
where $\mathbf{\Phi} \in \mathbb{R}^{N \times M} (N \ll M)$ is the dictionary matrix whose any $N$ columns are linearly independent, $\mathbf{Y} \in \mathbb{R}^{N \times L}$ is the measurement matrix consisting of $L$ measurement vectors, $\mathbf{X} \in \mathbb{R}^{M \times L}$ is the source matrix with each row representing a possible source, and $\mathbf{V}$ is the white Gaussian noise matrix with each entry satisfying $\mathbf{V}_{ij}\sim \mathcal{N}(0,\lambda)$. The key assumption under the MMV model is  that the support (i.e. locations of nonzero entries) of every column vector $\mathbf{X}_{\cdot i} \; (\forall i)$ \footnote{The $i$-th column of $\mathbf{X}$ is denoted by $\mathbf{X}_{\cdot i}$. The $i$-th row of $\mathbf{X}$ is denoted by $\mathbf{X}_{i\cdot}$ (also called  the $i$-th source).} is identical (referred as \emph{the common sparsity assumption} in the literature \cite{Cotter2005}). The MMV problem is often encountered in practical applications, such as neuroelectromagnetic source localization and direction-of-arrival estimation.

Most algorithms for the MMV problem can be roughly divided  into greedy methods, methods based on mixed norm optimization, iterative reweighted methods, and Bayesian methods.

Iterative reweighted methods have received attention because of their improved performance compared to their non-reweighted counterparts \cite{David2010reweighting,Candes2008reweighting}. In \cite{Candes2008reweighting},  an iterative reweighted $\ell_1$ minimization framework is employed. The framework can be directly used for the MMV problem and many MMV algorithms based on mixed norm optimization can be improved via the framework. On the other hand, iterative reweighted $\ell_2$ algorithms were also proposed \cite{David2010reweighting,Chartrand2008ICASSP}. The reweighted $\ell_2$ minimization framework for the MMV problem (in noisy case) computes the solution at the $(k+1)$-th iteration as follows \footnote{For convenience, we omit the superscript, $k$, on the right hand side of learning rules in the following.}:
\begin{eqnarray}
\mathbf{X}^{(k+1)} &=& \arg\min_\mathbf{x} \|\mathbf{Y} - \mathbf{\Phi} \mathbf{X}  \|_\mathcal{F}^2 + \lambda \sum_i w_i^{(k)} (\|\mathbf{X}_{i\cdot}\|_q)^2 \label{equ:framework_L2} \\
&=& \mathbf{W}^{(k)} \mathbf{\Phi}^T \big( \lambda \mathbf{I} + \mathbf{\Phi} \mathbf{W}^{(k)} \mathbf{\Phi}^T  \big)^{-1} \mathbf{Y} \label{equ:solution_L2framework}
\end{eqnarray}
where typically $q=2$, $\mathbf{W}^{(k)}$ is a diagonal weighting matrix at the $k$-th iteration with $i$-th diagonal element being $1/w_i^{(k)}$, and $w_i^{(k)}$ depends on the previous estimate of $\mathbf{X}$. Recently, Wipf et al \cite{David2010reweighting} unified most existing iterative reweighted algorithms as belonging to the family of \emph{separable} reweighted algorithms, whose weighting $w_i$ of a given row $\mathbf{X}_{i\cdot}$ at each iteration is only a function of that individual row from the previous iteration. Further, they proposed \emph{nonseparable} reweighted  algorithms via  variational approaches, which outperform many existing separable reweighted algorithms.

In our previous work \cite{Zhilin2010a,Zhilin2010b} we showed that temporal correlation in sources $\mathbf{X}_{i\cdot}$ seriously deteriorates recovery performance of existing algorithms and proposed a block sparse Bayesian learning (bSBL) framework, in which we  incorporated  temporal correlation and derived effective algorithms. These algorithms operate in the hyperparameter space, not in the source space as most sparse signal recovery algorithms do. Therefore, it is not clear what the connection of the bSBL framework is to other sparse signal recovery frameworks, such as  the reweighted $\ell_2$ in (\ref{equ:framework_L2}). In this work, based on the cost function in the bSBL framework, we derive an iterative reweighted $\ell_2$ SBL algorithm with superior performance, which directly operates in the source space. Furthermore, motivated by the intuition gained from the algorithm and  analytical insights, we propose a strategy to modify existing reweighted $\ell_2$ algorithms to incorporate temporal correlation of sources, and use two typical algorithms as illustrations. The strategy is shown to be effective.

\section{The Block Sparse Bayesian Learning Framework}
\label{sec:bSBL}

The block sparse Bayesian learning (bSBL) framework \cite{Zhilin2010a,Zhilin2010b}  transforms the MMV problem to a single measurement vector  problem. This makes the modeling of temporal correlation much easier. First, we assume the rows of $\mathbf{X}$ are mutually independent, and the density of each row $\mathbf{X}_{i\cdot}$ is multivariate Gaussian, given by
\begin{eqnarray}
p(\mathbf{X}_{i\cdot};\gamma_i,\mathbf{B}_i) \sim \mathcal{N}(\mathbf{0},\gamma_i \mathbf{B}_i), \quad  i=1,\cdots,M
\nonumber
\end{eqnarray}
where $\gamma_i$ is a nonnegative hyperparameter controlling the row sparsity of $\mathbf{X}$ as in the basic sparse Bayesian learning \cite{David2007IEEE,Tipping2001}. When $\gamma_i = 0$,  the associated $i$-th row of $\mathbf{X}$ becomes zero. $\mathbf{B}_i$ is an unknown positive definite correlation matrix.

By letting $\mathbf{y}=\mathrm{vec}(\mathbf{Y}^T) \in \mathbb{R}^{NL \times 1}$, $\mathbf{D} = \mathbf{\Phi} \otimes \mathbf{I}_L$ \footnote{We denote the $L \times L$ identity matrix  by $\mathbf{I}_L$. When the dimension is evident from the context, for simplicity we  use $\mathbf{I}$. $\otimes$ is the Kronecker product.}, $\mathbf{x}=\mathrm{vec}(\mathbf{X}^T) \in \mathbb{R}^{ML \times 1}$ and $\mathbf{v}=\mathrm{vec}(\mathbf{V}^T)$, where $\mathrm{vec}(\mathbf{A})$ denotes the vectorization of the matrix $\mathbf{A}$ formed by stacking its columns into a single column vector, we can transform the MMV model (\ref{equ:MMV basicmodel}) to the block single vector model as follows
\begin{eqnarray}
\mathbf{y}= \mathbf{D} \mathbf{x} + \mathbf{v}.
\label{equ:blocksparsemodel}
\end{eqnarray}
To elaborate on the block sparsity model (\ref{equ:blocksparsemodel}), we rewrite it as $\mathbf{y}= [\mathbf{\Phi}_1 \otimes \mathbf{I}_L, \cdots, \mathbf{\Phi}_M \otimes \mathbf{I}_L] [\mathbf{x}_1^T,\cdots,\mathbf{x}_M^T]^T + \mathbf{v} = \sum_{i=1}^M (\mathbf{\Phi}_i \otimes \mathbf{I}_L) \mathbf{x}_i + \mathbf{v}$, where $\mathbf{\Phi}_i$ is the $i$-th column of $\mathbf{\Phi}$, $\mathbf{x}_i \in \mathbb{R}^{L \times 1}$ is the $i$-th block in $\mathbf{x}$ and it is the transposed $i$-th row of $\mathbf{X}$ in the original MMV model (\ref{equ:MMV basicmodel}), i.e. $\mathbf{x}_i = \mathbf{X}_{i \cdot}^T$.  $K$ nonzero rows in $\mathbf{X}$ means $K$ nonzero blocks in $\mathbf{x}$. Thus we refer to $\mathbf{x}$ as block-sparse.

For the block model (\ref{equ:blocksparsemodel}), the Gaussian likelihood  is $p(\mathbf{y}|\mathbf{x}; \lambda) \sim \mathcal{N}_{y|x}(\mathbf{D}\mathbf{x}, \lambda \mathbf{I})$. The prior for $\mathbf{x}$ is given by
$p(\mathbf{x}; \gamma_i,\mathbf{B}_i,\forall i) \sim  \mathcal{N}_x(\textbf{0},\mathbf{\Sigma}_0) $, where  $\mathbf{\Sigma}_0$ is a block diagonal matrix with the $i$-th diagonal block $\gamma_i \mathbf{B}_i$ ($\forall i$).
Given the hyperparameters $\Theta \triangleq \{\lambda,\gamma_i,\mathbf{B}_i,\forall i\}$, the Maximum A Posterior (MAP) estimate of $\mathbf{x}$ can be directly obtained from the posterior of the model. To estimate these hyperparameters, we can use the Type-II maximum likelihood method \cite{Tipping2001}, which marginalizes over $\mathbf{x}$ and then performs maximum likelihood estimation, leading to the cost function:
\begin{eqnarray}
\mathcal{L}(\Theta) &\triangleq &  -2 \log \int p(\mathbf{y}|\mathbf{x};\lambda) p(\mathbf{x};\gamma_i,\mathbf{B}_i,\forall i) d \mathbf{x} \nonumber \\
&=& \log|\lambda \mathbf{I} + \mathbf{D} \mathbf{\Sigma}_0 \mathbf{D}^T  | + \mathbf{y}^T (\lambda \mathbf{I} + \mathbf{D} \mathbf{\Sigma}_0 \mathbf{D}^T)^{-1} \mathbf{y},
\label{equ:costfunc}
\end{eqnarray}
where $\gamma \triangleq [\gamma_1,\cdots,\gamma_M]^T$. We refer to the whole framework including the solution estimation of $\mathbf{x}$  and the hyperparameter estimation as the  bSBL framework. Note that in contrast to the original SBL framework, the bSBL framework models the temporal correlation structure of sources in the prior density via the matrix $\mathbf{B}_i\,(\forall i)$.

\section{Iterative Reweighted Sparse Bayesian Learning Algorithms}
\label{sec:Reweighting}

Based on the cost function (\ref{equ:costfunc}), we can derive efficient algorithms that exploit temporal correlation of sources \cite{Zhilin2010a,Zhilin2010b}. But these algorithms directly operate in the hyperparameter space (i.e. the $\gamma$-space). So, it is not clear what  their connection is to other sparse signal recovery algorithms that directly operate in the source space (i.e. the $\mathbf{X}$-space) by minimizing penalties on the sparsity of $\mathbf{X}$. Particularly, it is interesting to see if we can transplant the benefits gained from the bSBL framework to other sparse signal recovery frameworks such as the iterative reweighted $\ell_2$ minimization framework (\ref{equ:framework_L2}), improving algorithms belonging to those frameworks. Following the approach developed  by Wipf et al
\cite{David2010reweighting} for the single measurement vector problem, in the following we use  the duality theory \cite{BoydBook} to obtain a penalty in the source space, based on which we derive an iterative reweighted algorithm for the MMV problem.

\subsection{Algorithms}

First, we find that assigning a different covariance matrix $\mathbf{B}_i$ to each source $\mathbf{X}_{i\cdot}$ will result in overfitting in the learning of the hyperparameters. To overcome the overfitting, we simplify and consider using one matrix $\mathbf{B}$ to model all the source covariance matrixes. Thus $\mathbf{\Sigma}_0 = \mathbf{\Gamma} \otimes \mathbf{B}$ with $\mathbf{\Gamma} \triangleq \mathrm{diag}([\gamma_1,\cdots,\gamma_M])$. Simulations will show that this simplification leads to good results even if different sources have different temporal correlations (see Section \ref{sec:experiment}).

In order to transform the cost function (\ref{equ:costfunc}) to the source space, we use the identity: $\mathbf{y}^T (\lambda \mathbf{I} + \mathbf{D} \mathbf{\Sigma}_0 \mathbf{D}^T)^{-1} \mathbf{y} \equiv \min_\mathbf{x} \big[\frac{1}{\lambda} \|\mathbf{y}-\mathbf{Dx}\|_2^2 + \mathbf{x}^T \mathbf{\Sigma}_0^{-1} \mathbf{x}  \big]$, by which we can upper-bound  the cost function (\ref{equ:costfunc}) and obtain the bound
\begin{eqnarray}
\mathfrak{L}(\mathbf{x},\gamma,\mathbf{B}) = \log|\lambda \mathbf{I} + \mathbf{D} \mathbf{\Sigma}_0 \mathbf{D}^T  | + \frac{1}{\lambda} \|\mathbf{y}-\mathbf{Dx}\|_2^2 + \mathbf{x}^T \mathbf{\Sigma}_0^{-1} \mathbf{x} . \nonumber
\end{eqnarray}
By first minimizing over $\gamma$ and $\mathbf{B}$ and then minimizing over $\mathbf{x}$, we can get the cost function in the source space:
\begin{eqnarray}
\mathbf{x} = \arg\min_\mathbf{x} \|\mathbf{y}-\mathbf{Dx}\|_2^2 + \lambda g_{\mathrm{TC}}(\mathbf{x}),
\label{equ:x_space_expression}
\end{eqnarray}
where the penalty $g_{\mathrm{TC}}(\mathbf{x})$ is defined by
\begin{eqnarray}
g_{\mathrm{TC}}(\mathbf{x}) \triangleq \min_{\gamma \succeq \mathbf{0},\mathbf{B}\succ \mathbf{0}} \mathbf{x}^T \mathbf{\Sigma}_0^{-1} \mathbf{x}  + \log|\lambda \mathbf{I} + \mathbf{D} \mathbf{\Sigma}_0 \mathbf{D}^T  |.
\label{equ:g_definition}
\end{eqnarray}
From the definition (\ref{equ:g_definition}) we have
\begin{eqnarray}
g_{\mathrm{TC}}(\mathbf{x}) &\leq&  \mathbf{x}^T \mathbf{\Sigma}_0^{-1} \mathbf{x}  + \log|\lambda \mathbf{I} + \mathbf{D} \mathbf{\Sigma}_0 \mathbf{D}^T  | \nonumber \\
&=& \mathbf{x}^T \mathbf{\Sigma}_0^{-1} \mathbf{x} + \log|\mathbf{\Sigma}_0| + \log|\frac{1}{\lambda}\mathbf{D}^T \mathbf{D} + \mathbf{\Sigma}_0^{-1}| + NL \log \lambda \nonumber \\
& \leq & \mathbf{x}^T \mathbf{\Sigma}_0^{-1} \mathbf{x} + \log|\mathbf{\Sigma}_0| + \mathbf{z}^T \mathbf{\gamma}^{-1} - f^*(\mathbf{z}) + NL \log \lambda \nonumber
\end{eqnarray}
where in the last inequality we have used the conjugate relation
\begin{eqnarray}
\log \big| \frac{1}{\lambda}\mathbf{D}^T \mathbf{D} + \mathbf{\Sigma}_0^{-1} \big| = \min_{\mathbf{z}\succeq 0} \mathbf{z}^T \mathbf{\gamma}^{-1} - f^*(\mathbf{z}).
\label{equ:dual_relation}
\end{eqnarray}
Here we denote $\gamma^{-1} \triangleq [\gamma_1^{-1},\cdots,\gamma_M^{-1}]^T$, $\mathbf{z} \triangleq [z_1,\cdots,z_M]^T$, and  $f^*(\mathbf{z})$ is concave conjugate of $f(\gamma^{-1}) \triangleq \log|\frac{1}{\lambda}\mathbf{D}^T \mathbf{D} + \mathbf{\Sigma}_0^{-1}|$. Finally, reminding of $\mathbf{\Sigma}_0 = \mathbf{\Gamma} \otimes \mathbf{B}$, we have
\begin{eqnarray}
g_{\mathrm{TC}}(\mathbf{x}) \leq && NL \log \lambda - f^*(\mathbf{z}) + M\log |\mathbf{B}| + \nonumber \\
&&\sum_{i=1}^M \Big[ \frac{\mathbf{x}_i^T \mathbf{B}^{-1} \mathbf{x}_i + z_i}{\gamma_i} + L\log\gamma_i  \Big] . \label{equ:g_final}
\end{eqnarray}
Therefore, to solve the problem (\ref{equ:x_space_expression}) with (\ref{equ:g_final}), we can perform the coordinate descent method over $\mathbf{x},\mathbf{B},\mathbf{z}$ and $\gamma$, i.e,
\begin{eqnarray}
\min_{\mathbf{x},\mathbf{B},\mathbf{z}\succeq 0,\gamma \succeq 0} && \|\mathbf{y}-\mathbf{Dx}\|_2^2 + \lambda \Big[ \sum_{i=1}^M \Big( \frac{\mathbf{x}_i^T \mathbf{B}^{-1} \mathbf{x}_i + z_i}{\gamma_i}  \nonumber \\
&& +   L\log\gamma_i \Big) +   M\log |\mathbf{B}| - f^*(\mathbf{z})  \Big].
\label{equ:costfunc_new}
\end{eqnarray}
Compared to the framework (\ref{equ:framework_L2}), we can see $1/\gamma_i$ can be seen as the weighting for the corresponding $\mathbf{x}_i^T \mathbf{B}^{-1} \mathbf{x}_i$. But instead of applying $\ell_q$ norm on $\mathbf{x}_i$ (i.e. the $i$-th row of $\mathbf{X}$) as done in existing iterative reweighted $\ell_2$ algorithms, our algorithm computes $\mathbf{x}_i^T \mathbf{B}^{-1} \mathbf{x}_i$, i.e. the quadratic \emph{Mahalanobis distance} of $\mathbf{x}_i$ and its mean vector $\mathbf{0}$.

By minimizing (\ref{equ:costfunc_new}) over $\mathbf{x}$, the updating rule for $\mathbf{x}$ is given by
\begin{eqnarray}
\mathbf{x}^{(k+1)} = \mathbf{\Sigma}_0 \mathbf{D}^T (\lambda \mathbf{I} + \mathbf{D} \mathbf{\Sigma}_0 \mathbf{D}^T )^{-1} \mathbf{y}.
\label{equ:updating_x}
\end{eqnarray}
According to the dual property \cite{BoydBook}, from the relation (\ref{equ:dual_relation}), the optimal $\mathbf{z}$ is directly given by
\begin{eqnarray}
z_i &=& \frac{\partial \log |\frac{1}{\lambda}\mathbf{D}^T \mathbf{D} + \mathbf{\Sigma}_0^{-1}|}{\partial (\gamma_i^{-1})} \nonumber \\
&=& L\gamma_i - \gamma_i^2 \mathrm{Tr} \Big[\mathbf{B} \mathbf{D}_i^T \big( \lambda \mathbf{I} + \mathbf{D} \mathbf{\Sigma}_0 \mathbf{D}^T  \big)^{-1} \mathbf{D}_i   \Big], \; \forall i
\label{equ:updating_z}
\end{eqnarray}
where $\mathbf{D}_i$  consists of columns of $\mathbf{D}$ from the $((i-1)L+1)$-th to the $(iL)$-th. From (\ref{equ:costfunc_new}) the optimal $\gamma_i$ for fixed $\mathbf{x},\mathbf{z},\mathbf{B}$ is given by $\gamma_i = \frac{1}{L}[\mathbf{x}_i^T \mathbf{B}^{-1} \mathbf{x}_i + z_i]$. Substituting Eq.(\ref{equ:updating_z}) into it, we have
\begin{eqnarray}
\gamma_i^{(k+1)} &=& \frac{\mathbf{x}_i^T \mathbf{B}^{-1} \mathbf{x}_i}{L} + \gamma_i \nonumber \\
&& - \frac{\gamma_i^2}{L} \mathrm{Tr} \Big[\mathbf{B} \mathbf{D}_i^T \big( \lambda \mathbf{I} + \mathbf{D} \mathbf{\Sigma}_0 \mathbf{D}^T  \big)^{-1} \mathbf{D}_i   \Big], \; \forall i
\label{equ:updating_gamma}
\end{eqnarray}
By minimizing (\ref{equ:costfunc_new}) over $\mathbf{B}$, the updating rule for $\mathbf{B}$ is given by
\begin{eqnarray}
\mathbf{B}^{(k+1)} = \overline{\mathbf{B}}/\|\overline{\mathbf{B}}\|_\mathcal{F}, \quad \mathrm{with} \quad \overline{\mathbf{B}} =  \sum_{i=1}^M \frac{\mathbf{x}_i \mathbf{x}_i^T}{\gamma_i}.
\label{equ:updating_B}
\end{eqnarray}

The updating rules (\ref{equ:updating_x}) (\ref{equ:updating_gamma}) and (\ref{equ:updating_B}) are our  reweighted algorithm minimizing the penalty based on quadratic Mahalanobis distance of $\mathbf{x}_i$. Since for a given $i$, the weighting $1/\gamma_i$ depends on the whole estimated source matrix in the previous iteration (via $\mathbf{B}$ and $\mathbf{\Sigma}_0$), the algorithm is a \emph{nonseparable} reweighted algorithm.

The complexity of this algorithm is high because it learns the parameters in a higher dimensional space than the original problem space. For example, consider the bSBL framework, in which the dictionary matrix $\mathbf{D}$ is of the size $NL \times ML$, while in the original MMV model the dictionary matrix is of the size $N \times M$. We use an approximation to simplify the algorithm and develop an efficient variant. Using the approximation:
\begin{eqnarray}
\big(\lambda \mathbf{I}_{NL} +  \mathbf{D} \mathbf{\Sigma}_0 \mathbf{D}^T \big)^{-1}
\approx  \big(\lambda \mathbf{I}_N +  \mathbf{\Phi} \mathbf{\Gamma} \mathbf{\Phi}^T \big)^{-1} \otimes \mathbf{B}^{-1},
\label{equ:approximationFormula}
\end{eqnarray}
which takes the equal sign when $\lambda = 0$ or $\mathbf{B} = \mathbf{I}$, the updating rule  (\ref{equ:updating_x}) can be transformed to
\begin{eqnarray}
\mathbf{X}^{(k+1)} = \mathbf{W} \mathbf{\Phi}^T \big( \lambda \mathbf{I} + \mathbf{\Phi} \mathbf{W} \mathbf{\Phi}^T \big)^{-1} \mathbf{Y},
\label{equ:updating_x_original}
\end{eqnarray}
where $\mathbf{W} \triangleq \mathrm{diag}([1/w_1,\cdots,1/w_M])$ with $w_i \triangleq 1/\gamma_i$. Using the same approximation, the last term in (\ref{equ:updating_gamma}) becomes
\begin{eqnarray}
&& \mathrm{Tr} \Big[\mathbf{B} \mathbf{D}_i^T \big( \lambda \mathbf{I}_{NL} + \mathbf{D} \mathbf{\Sigma}_0 \mathbf{D}^T  \big)^{-1} \mathbf{D}_i   \Big] \nonumber \\
&& \approx \mathrm{Tr} \Big[\mathbf{B} (\mathbf{\Phi}_i^T \otimes \mathbf{I}) \big[ (\lambda \mathbf{I}_N + \mathbf{\Phi} \mathbf{W} \mathbf{\Phi}^T)^{-1} \otimes \mathbf{B}^{-1}  \big] (\mathbf{\Phi}_i \otimes \mathbf{I})   \Big] \nonumber \\
&& = L \mathbf{\Phi}_i^T(\lambda \mathbf{I}_N + \mathbf{\Phi} \mathbf{W} \mathbf{\Phi}^T)^{-1}\mathbf{\Phi}_i.  \nonumber
\end{eqnarray}
Therefore, from the updating rule of $\gamma_i$ (\ref{equ:updating_gamma}) we have
\begin{eqnarray}
w_i^{(k+1)}  =  \Big[\frac{1}{L} \mathbf{X}_{i\cdot} \mathbf{B}^{-1} \mathbf{X}_{i\cdot}^T   + \{( \mathbf{W}^{-1} + \frac{1}{\lambda} \mathbf{\Phi}^T \mathbf{\Phi} )^{-1}\}_{ii}\Big]^{-1}.
\label{equ:updating_gamma_original}
\end{eqnarray}
Accordingly, the updating rule for $\mathbf{B}$ becomes
\begin{eqnarray}
\mathbf{B}^{(k+1)} = \overline{\mathbf{B}}/\|\overline{\mathbf{B}}\|_\mathcal{F}, \quad \mathrm{with} \quad \overline{\mathbf{B}} =  \sum_{i=1}^M w_i  \mathbf{X}_{i \cdot}^T \mathbf{X}_{i \cdot}.
\label{equ:updating_B_original}
\end{eqnarray}
We denote the updating rules (\ref{equ:updating_x_original}) (\ref{equ:updating_gamma_original}) and (\ref{equ:updating_B_original}) by \emph{\textbf{ReSBL-QM}}. With the aid of singular value decomposition, the computational complexity of the algorithm is $\mathcal{O}(N^2 M)$ (The effect of $L$ can be removed by using the strategy in \cite{David2007IEEE}).

\subsection{Estimate the Regularization Parameter $\lambda$}

To estimate the regularization parameter $\lambda$, many methods have been proposed, such as  the modified L-curve method \cite{Cotter2005}. Here, straightforwardly following the Expectation-Maximization method in \cite{Zhilin2010a} and using the approximation  (\ref{equ:approximationFormula}), we derive a learning rule for $\lambda$, given by:
\begin{eqnarray}
\lambda^{(k+1)} = \frac{1}{NL} \|\mathbf{Y}-\mathbf{\Phi}\mathbf{X} \|_\mathcal{F}^2 + \frac{\lambda}{N} \mathrm{Tr}\big[ \mathbf{G} (  \lambda \mathbf{I} + \mathbf{G} )^{-1}   \big].
\nonumber
\end{eqnarray}
where  $\mathbf{G} \triangleq  \mathbf{\Phi} \mathbf{W} \mathbf{\Phi}^T$.

\subsection{Theoretical Analysis in the Noiseless Case}

For the noiseless inverse problem $\mathbf{Y} = \mathbf{\Phi}\mathbf{X}$, denote the generating sources by $\mathbf{X}_{\mathrm{gen}}$, which is the sparsest solution among all the possible solutions. Assume  $\mathbf{X}_{\mathrm{gen}}$ is full column-rank. Denote the true source number (i.e. the true number of nonzero rows in $\mathbf{X}_{\mathrm{gen}}$) by $K_0$. Now we have the following result on the global minimum of the cost function (\ref{equ:costfunc}):

\textbf{Theorem 1}  \emph{In the noiseless case, assuming $K_0 < (N+L)/2$, for the cost function (\ref{equ:costfunc}) the unique global minimum $\widehat{\gamma} = [\widehat{\gamma}_1,\cdots,\widehat{\gamma}_M]$ produces a source estimate $\widehat{\mathbf{X}}$  that equals to $\mathbf{X}_{\mathrm{gen}}$ irrespective of the estimated $\widehat{\mathbf{B}}_i, \, \forall i$, where $\widehat{\mathbf{X}}$ is obtained from $\mathrm{vec}(\widehat{\mathbf{X}}^T) = \widehat{\mathbf{x}} $ and $\widehat{\mathbf{x}}$ is computed using Eq.(\ref{equ:updating_x}).}

The proof is given in \cite{Zhilin2010b}. The theorem implies that even if the estimated $\widehat{\mathbf{B}}_i$ is different from the true $\mathbf{B}_i$, the estimated sources are the true sources at the global minimum of the cost function. Therefore the estimation error in $\widehat{\mathbf{B}}_i$ does not harm the recovery of true sources. As a reminder,  in deriving our algorithm, we assumed $\mathbf{B}_i = \mathbf{B}$ ($\forall i$) to avoid overfitting. The theorem ensures that this strategy does not harm the global minimum property.

In our  work \cite{Zhilin2010b} we have shown that $\mathbf{B}$ plays the role of whitening sources in the SBL procedure, which can be seen in  our algorithm as well. This gives us a motivation to improve some state-of-the-art reweighted $\ell_2$  algorithms by whitening the estimated sources in their weighting rules and penalties, detailed in the next section.

\section{Modify Existing Reweighted $\ell_2$ Methods}
\label{sec:extension}

Motivated by the above results and our analysis in \cite{Zhilin2010b}, we can modify many reweighted $\ell_2$ algorithms via replacing the $\ell_2$ norm  of $\mathbf{X}_{i\cdot}$ by some suitable function of its Mahalanobis distance. Note that similar modifications can be applied on reweighted $\ell_1$ algorithms.

The regularized M-FOCUSS \cite{Cotter2005} is a typical reweighted $\ell_2$ algorithm, which solves a reweighted $\ell_2$ minimization with weights $w_i^{(k)}=(\| \mathbf{X}_{i\cdot}^{(k)} \|_2^2)^{p/2-1}$ in each iteration. It is given by
\begin{eqnarray}
\mathbf{X}^{(k+1)} &=& \mathbf{W}^{(k)} \mathbf{\Phi}^T \big(\lambda \mathbf{I} + \mathbf{\Phi} \mathbf{W}^{(k)} \mathbf{\Phi}^T \big)^{-1} \mathbf{Y} \label{equ:mainIteration_FOCUSS} \\
 \mathbf{W}^{(k)} &=& \mathrm{diag}\{[1/w_1^{(k)},\cdots,1/w_M^{(k)}]  \} \nonumber \\
 w_i^{(k)} &=& \big( \| \mathbf{X}_{i\cdot}^{(k)} \|_2^2 \big)^{p/2-1}, \; p \in [0,2], \forall i \label{equ:weight_MFOCUSS}
\end{eqnarray}
We can modify the algorithm by changing (\ref{equ:weight_MFOCUSS}) to the following one:
\begin{eqnarray}
w_i^{(k)} &=& \big( \mathbf{X}^{(k)}_{i\cdot} (\mathbf{B}^{(k)})^{-1} (\mathbf{X}^{(k)}_{i\cdot})^T \big)^{p/2-1}, \; p \in [0,2], \forall i \label{equ:weight_tMFOCUSS}
\end{eqnarray}
The matrix $\mathbf{B}$ can be calculated using the learning rule (\ref{equ:updating_B_original}). We denote the modified algorithm by \emph{\textbf{tMFOCUSS}}.

In \cite{Chartrand2008ICASSP} Chartrand and Yin  proposed an  iterative reweighted $\ell_2$ algorithm based on the classic FOCUSS algorithm. Its MMV extension (denoted by \emph{\textbf{Iter-L2}}) changed (\ref{equ:weight_MFOCUSS}) to:
\begin{eqnarray}
w_i^{(k)} &=& \big( \| \mathbf{X}_{i\cdot}^{(k)} \|_2^2 + \epsilon \big)^{p/2-1}, \; p \in [0,2], \forall i \label{equ:weight_iterL2}
\end{eqnarray}
Their algorithm adopts the strategy: initially use a relatively large $\epsilon$, then repeating the process of decreasing $\epsilon$ after convergence and repeating the iteration (\ref{equ:mainIteration_FOCUSS}), dramatically improving the recovery ability. Similarly, we can modify the weighting (\ref{equ:weight_iterL2}) to the following rule incorporating the temporal correlation of sources:
\begin{eqnarray}
w_i^{(k)} &=& \big( \mathbf{X}^{(k)}_{i\cdot} (\mathbf{B}^{(k)})^{-1} (\mathbf{X}^{(k)}_{i\cdot})^T + \epsilon \big)^{p/2-1}, \; p \in [0,2], \forall i \label{equ:weight_tempIter2}
\end{eqnarray}
and adopts the same $\epsilon$-decreasing strategy. $\mathbf{B}$ is also given by (\ref{equ:updating_B_original}). We denote the modified algorithm by \emph{\textbf{tIter-L2}}.

The proposed tMFOCUSS and tIter-L2 have  convergence properties similar to   M-FOCUSS and Iter-L2, respectively. Due to space limit we omit theoretical analysis, and instead,  provide some representative simulation results in the next section.

\section{Experiments}
\label{sec:experiment}

In our experiments, a dictionary matrix $\mathbf{\Phi} \in \mathbb{R}^{N \times M}$ was created with columns uniformly drawing from the surface of a unit hypersphere. The source matrix $\mathbf{X}_{\mathrm{gen}} \in \mathbb{R}^{M \times L}$ was randomly generated with $K$ nonzero rows of unit norms, whose row locations were randomly chosen. Amplitudes of the $i$-th nonzero row were generated as an AR(1) process whose AR coefficient was denoted by $\beta_i$ \footnote{Since in our experiments the measurement vector number is very small ($L=3$ or 4), generating sources as AR(1) with various AR coefficient values is sufficient.}. Thus $\beta_i$ indicates the temporal correlation of the $i$-th source. The measurement matrix was constructed by $\mathbf{Y} = \mathbf{\Phi} \mathbf{X}_{\mathrm{gen}} + \mathbf{V}$, where $\mathbf{V}$ was a zero-mean homoscedastic Gaussian noise matrix with variance adjusted to have a desired value of $\mathrm{SNR}$. For each different experiment setting, we repeated 500 trials and averaged results. The performance measurement was the \emph{Failure Rate } defined in \cite{David2007IEEE}, which indicated the percentage of failed trials in the 500 trials. When noise was present, since we could not expect any algorithm to recover $\mathbf{X}_{\mathrm{gen}}$ exactly, we classified a trial as a failure trial if the $K$ largest estimated row-norms did not align with the support of $\mathbf{X}_{\mathrm{gen}}$. The compared algorithms included our proposed ReSBL-QM, tMFOCUSS, tIter-L2, the reweighted $\ell_2$ SBL in  \cite{David2010reweighting} (denoted by ReSBL-L2), M-FOCUSS \cite{Cotter2005}, Iter-L2 presented in Section \ref{sec:extension}, and Candes' reweighted $\ell_1$ algorithm \cite{Candes2008reweighting} (extended to the MMV case as suggested by \cite{David2010reweighting}, denoted by Iter-L1). For tMFOCUSS, M-FOCUSS, and Iter-L2, we set $p=0.8$, which gave the best performance in our simulations. For Iter-L1, we used 5 iterations.

\begin{figure}[b]
\begin{minipage}[b]{.48\linewidth}
  \centering
  \centerline{\epsfig{figure=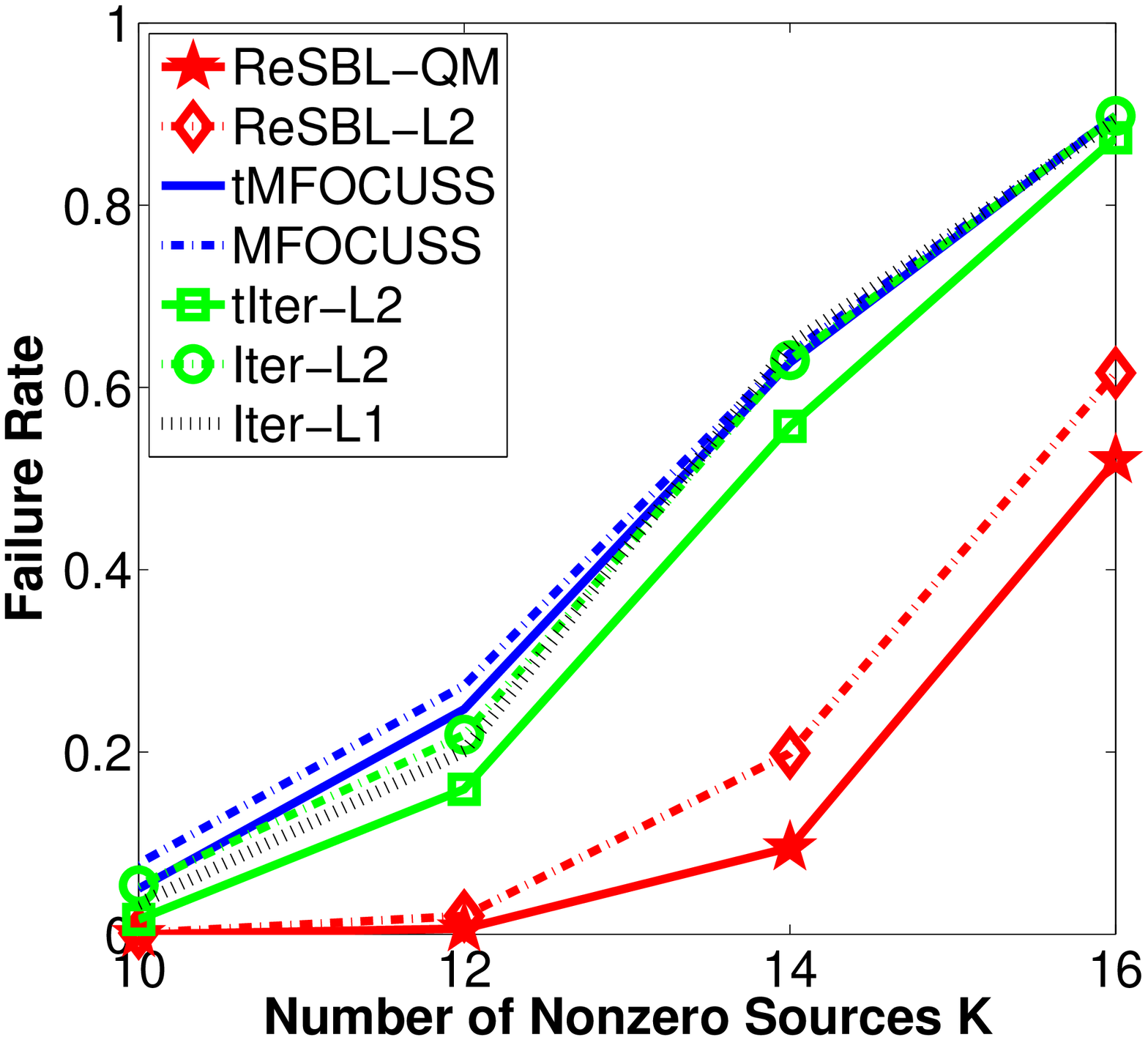,width=4.5cm,height=3.4cm}}
  \centerline{\footnotesize{(a) Low Correlation Case}}
\end{minipage}
\hfill
\begin{minipage}[b]{0.48\linewidth}
  \centering
  \centerline{\epsfig{figure=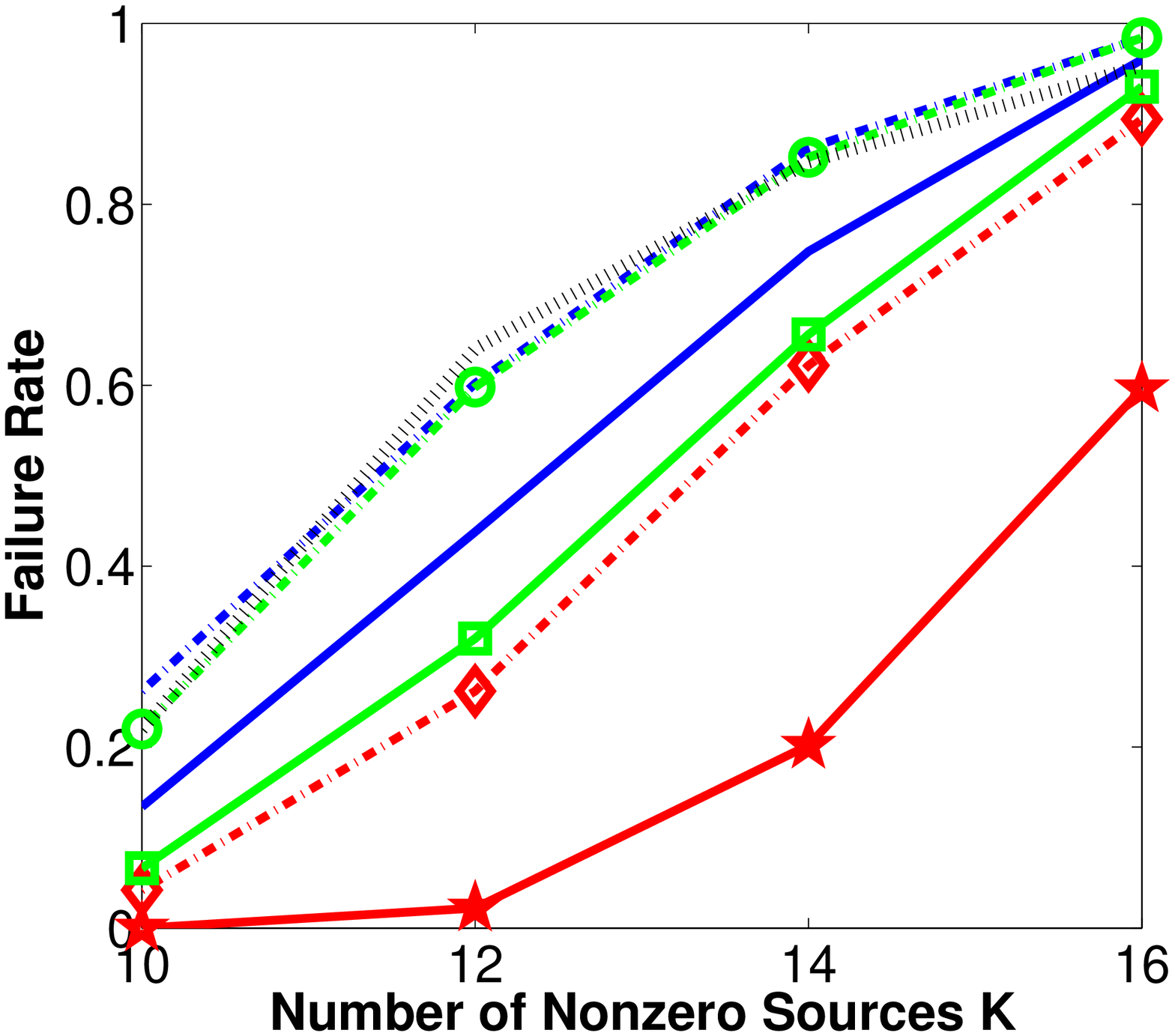,width=4.5cm,height=3.4cm}}
  \centerline{\footnotesize{(b) High Correlation Case}}
\end{minipage}
\caption{Performance when the nonzero source number changes.}
\label{fig:varyK}
\end{figure}

In the first experiment we fixed $N=25$, $M=100$,  $L=3$ and $\mathrm{SNR} = 25\mathrm{dB}$. The number of nonzero sources $K$ varied from 10 to 16. Fig.\ref{fig:varyK} (a) shows the results when each $\beta_i$ was uniformly chosen from $[0,0.5)$ at random. Fig.\ref{fig:varyK} (b) shows the results when each $\beta_i$ was uniformly chosen from $[0.5,1)$ at random.

\begin{figure}[h]
\begin{minipage}[b]{.48\linewidth}
  \centering
  \centerline{\epsfig{figure=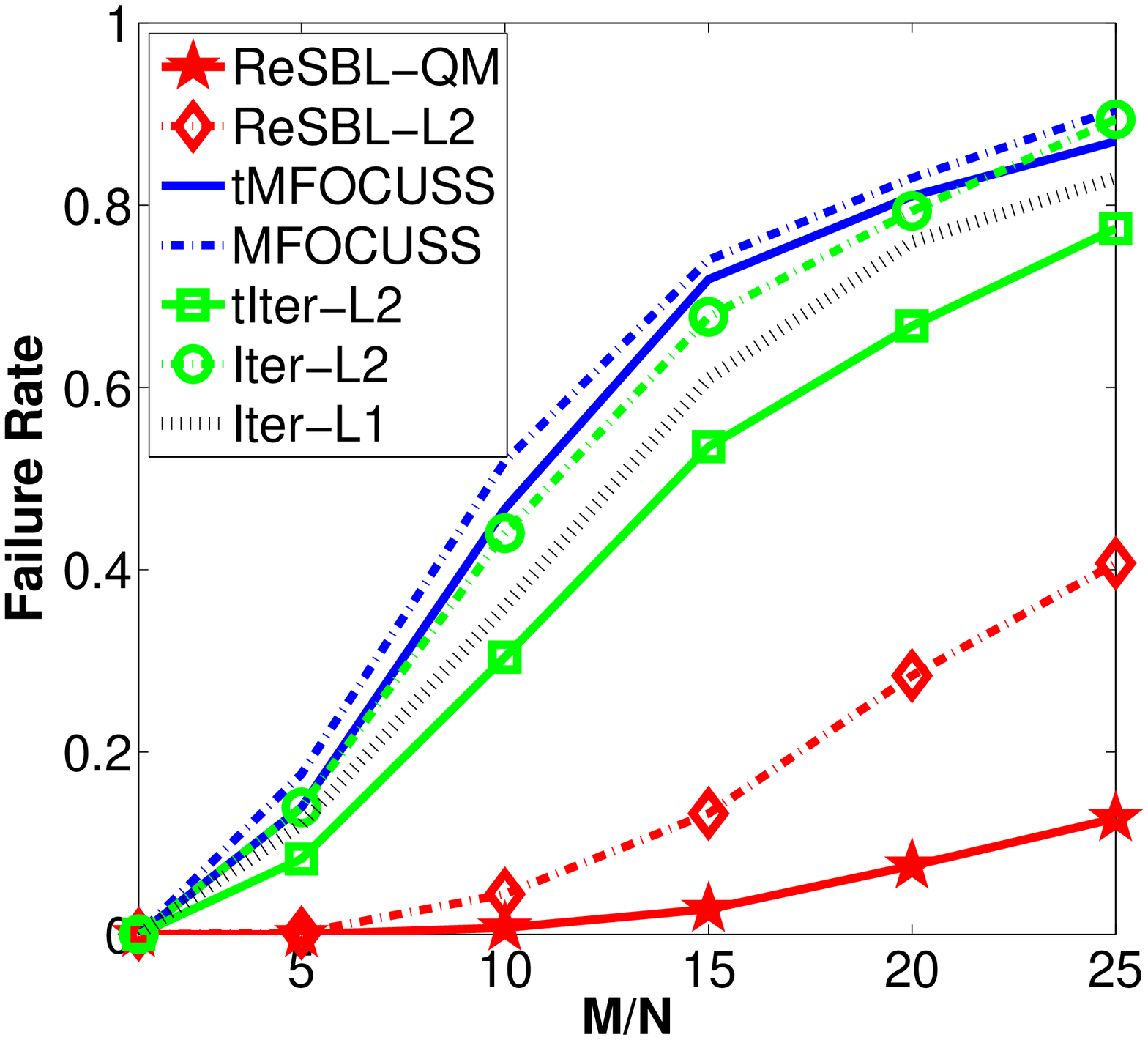,width=4.5cm,height=3.4cm}}
  \centerline{\footnotesize{(a) Low Correlation Case}}
\end{minipage}
\hfill
\begin{minipage}[b]{0.48\linewidth}
  \centering
  \centerline{\epsfig{figure=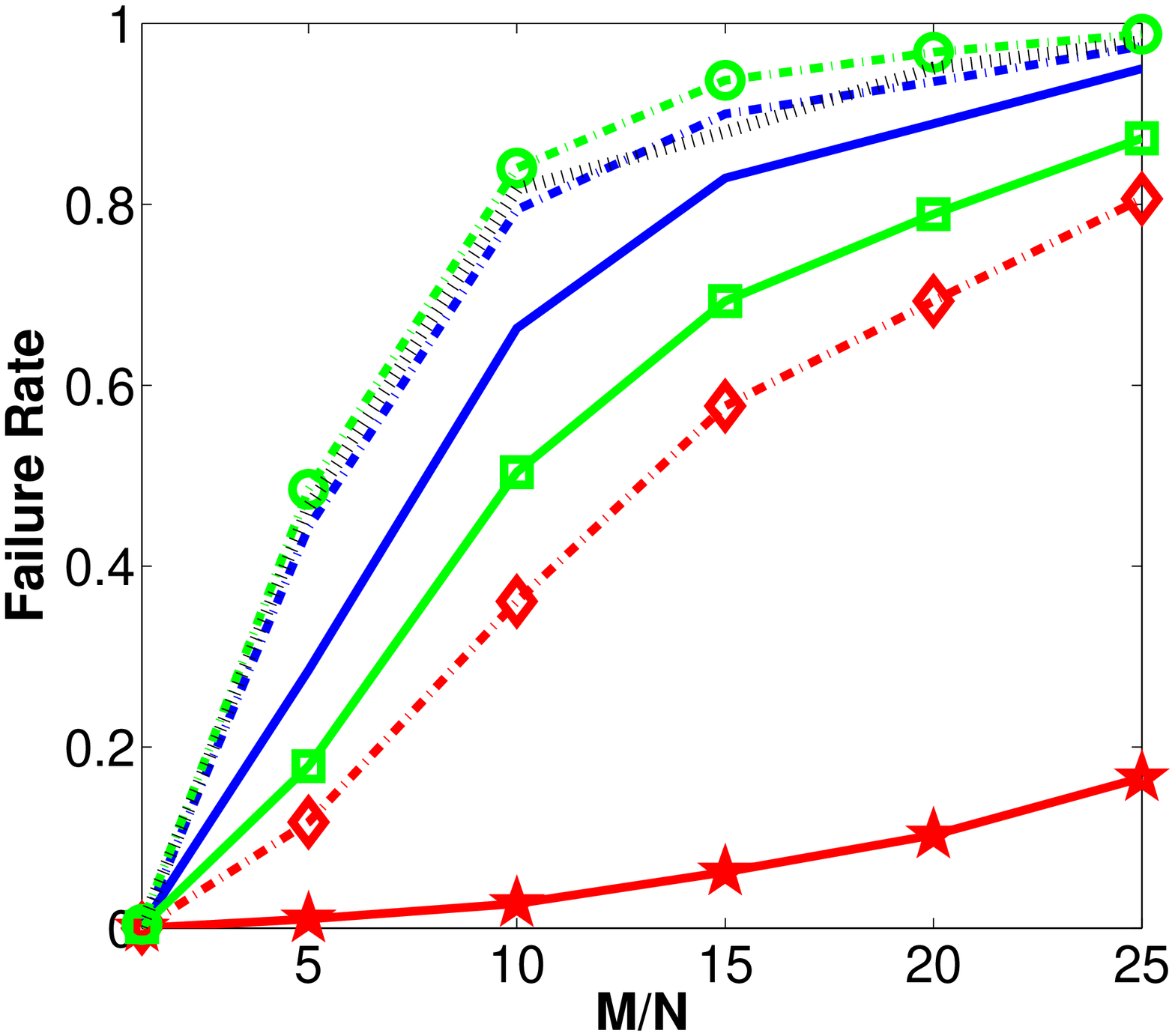,width=4.5cm,height=3.4cm}}
  \centerline{\footnotesize{(b) High Correlation Case}}
\end{minipage}
\caption{Performance when $M/N$ changes.}
\label{fig:varyMN}
\end{figure}

In the second experiment we fixed  $N = 25$, $L=4$, $K=12$, and $\mathrm{SNR} = 25\mathrm{dB}$, while $M/N$ varied from 1 to 25. $\beta_i$ ($\forall i$) in Fig.\ref{fig:varyMN} (a) and (b) were generated as in Fig.\ref{fig:varyK} (a) and (b), respectively. This experiment aims to see algorithms' performance in highly underdetermined inverse problems, which met in some applications such as neuroelectromagnetic source localization.

From the two experiments we can see that: (a) in all cases, the proposed ReSBL-QM has superior performance to other algorithms, capable to recover more sources and solve more highly underdetermined inverse problems; (b) without considering temporal correlation of sources, existing algorithms' performance significantly degrades with increasing correlation; (c) after incorporating the temporal structures of sources, the modified algorithms, i.e. tMFOCUSS and tIter-L2, have better performance than the original M-FOCUSS and Iter-L2, respectively. Also, we noted that our proposed algorithms are more effective when the norms of sources have no large difference (results are not shown here due to space limit).




\section{Conclusions}

In this paper, we derived an iterative reweighted sparse Bayesian algorithm exploiting the temporal structure of sources. Its simplified variant was also obtained, which has less computational load. Motivated by our analysis we modified some state-of-the-art reweighted $\ell_2$ algorithms achieving improved performance. This work not only provides some effective reweighted algorithms, but also provides a strategy to design effective reweighted algorithms enriching current algorithms on this topic.

\bibliographystyle{IEEEtran}

\bibliography{IEEEabrv,bookbibfile,sparsebibfile}

\end{document}